\documentclass[10pt,twocolumn]{article}
\usepackage{times}
\usepackage{graphicx}
\newtheorem{definition}{Definition}
\usepackage{times}
\usepackage{graphicx}
\usepackage{amsfonts}
\usepackage{amssymb}
\newenvironment{theorem}{\par\bgroup{\bf Theorem.}\it}{\egroup\par}

\ifx\pdfobj\undefined

\else

\fi

\def\R{{\mathbb R}}

\def\p{\par}

\def\paroff{\catcode`\^^M=10}

\iffalse

\else

\fi

\def\val#1{\left\lfloor{#1}\right\rfloor}

\begin{document}
\paroff

\date{
November 2003\footnote{
This TR was originally written in 2003, but submitted to Arxiv in 2007.
References have not been updated.
}
}

\title{\Large\bf Learning Similarity for Character Recognition and 3D Object Recognition}

\author{
Thomas M. Breuel\\
PARC, Inc.\\
3333 Coyote Hill Rd.\\
Palo Alto, CA 94304, USA
}

\maketitle

\section*{\centering Abstract}

{\em I describe an approach to similarity motivated
by Bayesian methods.  This yields a similarity function
that is learnable using a standard Bayesian methods.
The relationship of the approach to variable kernel
and variable metric methods is discussed.
The approach is related to variable kernel
Experimental results on character recognition and 3D object
recognition are presented.  }

\section{Introduction}

Visual object recognition, character recognition, speech recognition,
and a wide variety of statistical and engineering problems involve
classification.

Classifiers attempt to assign class labels to novel,
unlabeled data based on previously seen labeled training data.

For example, determining the identity of a letter (the ``class'') from
a scanned image of the latter (the ``feature vector'') is an example
of a classification problem occurring in optical character recognition
(OCR).

\p

Two very common approaches to solving classification problems are
Bayesian methods and nearest neighbor methods.

In Bayesian methods, we model class conditional distributions and use
those estimates for finding minimum error rate discriminant functions.

In nearest neighbor methods, we classify unknown feature vectors based
on their proximity in feature space (usually, some Euclidean space,
$\R^d$) to previously classified samples.

\p

Nearest neighbor methods can actually be viewed as a special case of
Bayesian methods if we view the nearest neighbor procedure as
implicitly using a non-parametric approximation of class conditional
densities.

Asymptotically, the error rate of nearest neighbor procedures
is known to be within a factor of two of the Bayes optimal
error rate \cite{Duda01}.

Just as important as the asymptotic error rate is how
quickly the error rate of a classifier decreases with increasing
amounts of training data.

\p

To achieve improvements in these areas, a number of authors (e.g.,
\cite{Lowe95,Friedman94,HasTib96,shankar00weight,DomPenGun00})
have proposed using similarity functions other than the Euclidean
distance in nearest neighbor classification, and give on-line or
off-line procedures for computing such similarity
functions\footnote{They are often referred to as ``adaptive similarity
metrics'', but they do not satisfy the metric axioms and to avoid
confusion, we refer to them here as ``similarity functions''.}

Another recent development is the increased demand in applications for
sound ways of determining the ``similarity'' of two objects in areas
like 3D visual object recognition, biometric identification, case
based reasoning, and information retrieval (e.g.,
\cite{faltings97probabilistic,HofPuz98}).

\p

This paper describes a notion of similarity that is directly grounded
in Bayesian statistics and that is learnable based on training
examples using a wide variety of well-know density estimation methods
and classifiers.

It then discusses the relationship between such a notion of
statistical similarity and nearest neighbor classification.

The approach is motivated with several examples.

Experiments on learning character recognition and 3D object
recognition are discussed.

\section{Some Bayesian Decision Theory}

To establish notation and background, let us briefly review a
few aspects of Bayesian decision theory relevant to classification
problems.

Bayesian decision\cite{Berger80,Duda01} tells us that the approach
for finding minimum error rate solutions to classification
problems is the following.\footnote{
Without loss of generality, we consider minimization of the
expected loss under a zero-one loss function only in this paper.
}

Let $\Omega$ be a finite set of possible classes.

Let our feature vectors $x$ be vectors in $\R^d$.

First, estimate the class conditional densities $P(\omega|x)$.

Then, choose the class $\omega\in\Omega$ that has the maximum posterior
probability given the input data $x\in\R^d$.

\p

The differences among different classification methods come down
to different tradeoffs and approaches in estimating $P(\omega|x)$.

$P(\omega|x)$ is usually estimated from a large set of training
samples $\{(x_1,\omega_1),\ldots,(x_n,\omega_n)\}$, the
training set.

Here, the $x_i$ are measurements or feature vectors, and
the $\omega_i$ are the corresponding classes.

\p

One of the most common ways of estimating $P(\omega|x)$ is
to estimate $P(x|\omega)$ and then apply Bayes rule:

\begin{equation}
P(\omega|x) = \frac{P(x|\omega)\,P(\omega)}{P(x)}
\end{equation}

\p

For example, if samples $x$ are generated by picking a per-class
prototype $x_\omega$ and adding Gaussian random noise $N \sim
G(0,\Sigma)$ to it, then $x \sim x_\omega + N$ or, equivalently,
$P(x|\omega) = G(x_\omega,\Sigma)$.

This may be extended to allowing multiple prototypes per class, giving
mixture of Gaussian models $P(x|\omega) = \sum_i
G(x_{\omega,i},\Sigma)$.

Such parametric models for the basis of many applications of
classification in control theory and speech recognition.

They are attractive because we can often derive the distribution
of the noise from first principles and estimate the parameters
of the noise distribution using closed-form approaches.

\p

Another approach to modeling $P(x|\omega)$ is that of many-parameter
or non-parametric density estimation, using, for example, multi-layer
perceptrons, logistic regression, Parzen windows, and many other
techniques.

In essence, this is a special case of a function interpolation problem,
where $P(x|\omega)$ is to be interpolated based on training samples.

In fact, in many cases, estimating $P(x|\omega)$ can be solved by
least square linear regression on the data set
$\{(x_1,y_i),\ldots,(x_n,y_n)\}$, where as the regression variable
$y_i$, we pick the value of the indicator function $y_i =
\val{\omega=\omega_i}$, that is, we set $y_i$ to $1$ if $\omega_i =
\omega$ and $0$ otherwise.

Logistic regression and classification using multi-layer perceptrons
are closely related to such an approach.

\p

Since, for classification under a given loss function, we are
only interested in $\arg\max_\omega P(\omega|x)$, many approximations
to the posterior density are equivalent from the point of
view of classification.

This can be expressed by saying that instead of estimating densities,
we attempt to find decision functions $D_\omega(x)$ such that
classifying according to $\omega(x) = \arg\max D_\omega(x)$ results in
minimum error rates.

Such an approach is taken by, for example, linear discriminant
analysis and support vector machines (it has been argued that this
relaxation of the density approximation problem results in lower error
rates).

\section{Bayesian Similarity and Classification}

The motivation for the Bayesian similarity model
introduced in this paper is the following.

Assume we are performing nearest neighbor classification.

We are given a prototype $x'$ together with its class label $\omega'$
and an unknown vector $x$ to be classified.

If we could estimate the probability that $x$ and $x'$ represent
the same class, then we could use this to determine the probability
that vector $x$ comes from class $\omega'$.

\p

Let us write this probability as $P(S|x,x')$, where $S$ is a binary
variable, $S=1$ is $x$ and $x'$ come from the same class, and $S=0$
otherwise.

We can express $P(S|x,x')$ in terms of $P(\omega|x)$ and
$P(\omega|x')$ and use this as the definition of Bayesian
statistical similarity.

\begin{definition}
{\em The (Bayesian) statistical similarity function} $S(x,x')$ is the
conditional distribution $P(S|x,x')$, where
\begin{equation}
S(x,x') = P(S=1|x,x') = \sum_{\omega\in\Omega} P(\omega|x) P(\omega|x')  \label{eqsame}
\end{equation}
\end{definition}

Note that in this definition, the distributions $P(\omega,x)$ and
$P(\omega,x')$ need not be the same.

\p

There are a number of properties we should observe.

First, statistical similarity functions assume values in the
interval $[0,1]$.

Also, statistical similarity is dependent to some degree on the
classification problem we are considering (although we will see below
that statistical similarity can generalize to a wider variety of
classification problems than, say, a set of discriminant functions).

A value of $1$ means that two feature vectors $x$ and $x'$ are
known to be in the same class.

However, $S(x,x)$ can be less than one, namely when the feature vector
$x$ cannot be classified unambiguously.

\p

Now that we have a formal expression for $P(S|x,x')$, let us
look at the classification rule.

For this, we first need another definition.

\begin{definition}
Given some $\omega\in\Omega$, let us call $x_\omega$ an {\em unambiguous
exemplar} for class $\omega$ iff $P(\omega|x_\omega) = 1$; because of
normalization, this also means that $P(\omega'|x_\omega) = 0$ when
$\omega'\neq\omega$, or $P(\omega'|x_\omega) =
\delta(\omega',\omega)$.
\end{definition}

If $x_0$ is an unambiguous exemplar for class $\omega_0$, then

\begin{eqnarray}
P(S=1|x,x_0) &=& \sum_\omega P(\omega|x) P(\omega|x_0) \\
	     &=& \sum_\omega P(\omega|x) \delta(\omega,\omega_0)\\
	     &=& P(\omega|x)
\end{eqnarray}

Therefore, we have shown the following:

\begin{theorem}
If $x_0$ is an unambiguous exemplar for class $\omega_0$, then
$P(\omega_0|x) = P(S=1|x,x_0)$.
\end{theorem}

\p

The point of these derivations was to make a connection between
statistical similarity functions and Bayesian decision theory.

Overall, what this shows is that we can represent $P(\omega|x)$
as a statistical similarity function $P(S=1|x,x')$ and a set
of unambiguous exemplars $\{x_\omega\}$.

It gives us a prescription for constructing a nearest neighbor
classifier for many kinds of classification problems that is
guaranteed to achieve the Bayes optimal error rate.

\p

Of course, not all classification problems have unambiguous exemplars;
an analysis of such cases goes beyond the scope of this paper, and it
is probably not necessary for real-world applications.

For actual applications, we can use methods of machine learning for
estimating the statistical similarity function and then pick a set of
exemplars that empirically minimizes misclassification rate in a way
analogous to other nearest neighbor methods.

\section{Motivation}

Now that we have introduced statistical similarity functions,
we might ask what advantages they could have over either
models of posterior distributions or nearest neighbor methods.

The use of statistical similarity functions is somewhat analogous to
Bayes rule: we apply Bayes rule when we find the estimation of
$P(x|\omega)$ more convenient than the estimation of $P(\omega|x)$.

In fact, there are several important ways in which the estimation of
$P(S|x,x')$ is more convenient than estimating class conditional or
posterior distributions.

\p

First, and perhaps most importantly, learning $P(S|x,x')$ can be
done with unlabeled training data in some important cases.

In 3D visual object recognition, we have a wealth of
unlabeled training data available in the form of motion sequences.

These motion sequences give us different appearances of the same
object in successive frames and can be used to train $P(S|x,x')$
(see \cite{EdeWei91} for an example of a system that takes advantage
of this).

Furthermore, we would expect statistical similarity to be able
to take advantage of some properties that are independent of
object class; we will return to this point in the next section.

Also, we often have to solve a set of related classification
problems, for which we keep $P(S|x,x_\omega)$ constant but
use different sets of prototypes $x_\omega$.

\p

The above definitions and theorems are intended to motivate the use of
statistical similarity and to make a connection with Bayesian
approaches to classification based on class conditional and posterior
densities.

However, having a statistical similarity measure available does open
up new applications that do not fit well into a traditional
classification framework.

For example, in a case-based reasoning framework, information
retrieval, or 3D visual object recognition framework, we may
not have a meaningful set of {\it a priori} classifications.

Rather, the goal of the problem is to find the case, text, or
view that is most likely to have been derived from the same
underlying situation as the query.

In fact, several authors have formulated specific statistical models
for statistical similarity in case-based reasoning
\cite{faltings97probabilistic} and
information retrieval (e.g., \cite{HofPuz98,Hofmann1999:plsi}).

\section{An Example}

Consider a classification problem in which the observed vectors are
distributed according to $x \sim x_\omega + N$, where $x_\omega$ is a
class prototype and $N$ is iid noise, independent of the object class.

Then, $P(x|\omega) = N(x-x_\omega)$.

Since $P(S|x,x_\omega) = P(\omega|x) =
\frac{P(X|\omega)P(\omega)}{P(\omega)} =
\frac{N(x-x_\omega)P(\omega)}{\sum_{\omega'}N(x-x_{\omega'})}$, we see
that $P(S|x,x_\omega)$ is translation invariant: if we translate $x$
and the prototypes $x_\omega$, classification will be carried out the
same way.

Furthermore, staying with this example, if the prototypes $x_\omega$
are displaced by different amounts $\Delta_\omega$,
$P(S|x,x_\omega+\Delta_\omega)$ may not be an accurate estimate of
$P(\omega|x)$ anymore, but the decision rule $\arg\max_\omega
P(S|x,x_\omega+\Delta_\omega)$ can still be seen to be correct.

In practice, $N$ may not be completely independent of $x$, but
if it varies slowly, we can choose models of $P(S|x,x')$
that take advantage of this fact.

\p

In fact, this last example provides a connection with adaptive
metric models.

Consider a simple adaptive metric model in which we optimize a
quadratic form $Q$ for our metric in order to minimize the error rate;
that is, we use as our decision rule
\begin{equation}
\omega(x) = \arg\min_\omega (x-x_\omega)\cdot Q \cdot(x-x_\omega) \label{optq}
\end{equation}

If our decision rule is $\hat{\omega}(x) = \arg\max_\omega
P(S|x,x_\omega)$ and our noise model $N$ is a Gaussian $G(0,\Sigma)$,
then, by the above argument,
\begin{eqnarray}
\omega(x) &=& \arg\max_\omega P(S|x,x_\omega)\\
&=& \arg\max_\omega \frac{P(\omega)}{P(x)} G(x-x_\omega,\Sigma)\\
&=& \arg\max_\omega G(x-x_\omega,\Sigma)\\
&=& \arg\max_\omega e^{\frac{-1}{2||\Sigma||}(x-x_\omega)\cdot\Sigma\cdot(x-x_\omega)}\\
&=& \arg\min_\omega (x-x_\omega)\cdot\Sigma\cdot(x-x_\omega) \label{optsim}
\end{eqnarray}

By comparing Equation~\ref{optsim} and Equation~\ref{optq}, we see
that we can use $\Sigma$ as the quadratic form $Q$ (the choice is not
entirely unique).

\section{Character Recognition}

The above ideas were tested on an isolated handwritten character recognition
task using the NIST 3 database \cite{NIST-3} (see also \cite{LeCun98}
for a state-of-the-art character recognition system and comparisons
of a large number of classifiers).

Similar experiments have been used in other works on variable and
adaptive metric methods (e.g., \cite{DomPenGun00}).

\p

The overall idea is to estimate $P(S|x,x')$ using multi-layer
perceptrons (MLPs) as a simple and well-studied trainable model
of posterior probabilities.

Then, we use $P(S|x,x')$ as our ``distance'' in a $k$-nearest neighbor
classifier and compare its performance with the performance of a
standard nearest neighbor classifier.

\p

The images used in these experiments were images of handwritten digits
from the NIST 3 database.

Randomly selected images from the first 1000 writers were used for all
training, randomly selected images from a set of 200 separate writers
were used for testing.

For feature extraction, bounding boxes for characters were computed and
the characters were rescaled uniformly to fit into a $40\times 40$
image.

The resulting character image was slant corrected based on its second
order moments.

The uncorrected and slant corrected images form the first two feature
maps.

Derivatives were estimated along multiples of $\frac{\pi}{5}$ degrees,
resulting in five feature maps.

Additionally, feature maps of interior regions, skeletal endpoints, and
skeletal junction points were computed.

Each of the resulting feature maps was anti-aliased and scaled down to
a $10\times 10$ grid.

This results in 10 $10\times 10$ feature maps, or a 1000 dimensional
feature vector.

(Experiments were also carried out with subsets of these feature maps
consisting of only the raw image, 100 dimensional, or the raw image,
the slant corrected image, and derivatives, 700 dimensional, with
similar results.)

This feature extraction method was chosen because it has worked well
for character recognition using multi-layer perceptrons as classifiers
\cite{breuel-ocr};

however, there is no reason to believe that it is a particularly good
representation for the purpose of learning statistical similarity functions,
and the performance of the system can probably be improved by experimenting
with other feature extraction methods.

\p

To obtain statistical similarity models a multi-layer perceptron (MLP)
was trained using gradient descent training.

It has been shown (see \cite{Duda01} p.304) that training a
multi-layer perceptron under a least square error criterion and binary
output variables results in an approximation to the posterior
probability distribution.

The feature vectors from each image were concatenated to yield
the feature vector that formed the input to the MLP.

When the classes corresponding to the feature vectors in the NIST
database were the same, the target output during training was set
to $1$, otherwise $0$.

\p

After estimating a statistical similarity function this way,
the statistical similarity function was used in a simple
nearest neighbor classifier.

To select the prototypes for the nearest neighbor classifier, feature
vectors from the training set were compared to the set of prototypes
(initially empty) and the class associated with the most similar,
according to the statistical similarity function, was returned as the
classification.

Whenever the classification was incorrect, the incorrectly classified
feature vector was added to the set of prototypes.

This process was stopped when the set of prototypes had grown to
200 prototypes.

\p

To estimate misclassification rates, 5000 feature vectors were
selected from a separate test set and classified like the training
vectors (however, misclassified feature vectors were not added during
the set of prototypes).

As a control, the same training and testing process was carried out
using Euclidean distance.

The results of these experiments are shown in Table~\ref{tab-res1}.

They show a 2.7-fold improvement of using statistical similarity
over Euclidean distance.

\begin{table}[t]
\begin{center}
\begin{tabular}{|r|r|} \hline
Statistical Similarity  & Euclidean       \\
Nearest Neighbor        & Nearest Neighbor\\ \hline
2.6\%                   & 9.5\%           \\ \hline
\end{tabular}
\end{center}
\caption{\label{tab-res1}
An experimental comparison of the performance of Euclidean nearest
neighbor methods with statistical similarity based nearest neighbor
methods.  The error rates are derived from 5000 test samples, using
200 prototypes selected as described in the paper.
}
\end{table}

\p

In a second set of experiments, the statistical similarity function
was trained not on randomly selected pairs of feature vectors, but
only on pairs of feature vectors from the same writer.

This means that the statistical similarity function characterizes
the variability for individual writers.

For testing, feature vectors from 200 writers not in the training
set were used.

For each writer, the first instance of each character was used
as a prototype, resulting in 10 prototypes per writer.

These prototypes were then used to classify the remaining samples
from the same writer.

These results are shown in Table~\ref{tab-res2}.

The results show a 4.4-fold improvement of statistical similarity
over Euclidean nearest neighbor methods.

\begin{table}[t]
\begin{center}
\begin{tabular}{|r|r|} \hline
Statistical Similarity   & Euclidean        \\
Nearest Neighbor         & Nearest Neighbor \\ \hline
5.1\%                    & 22.6\%           \\ \hline
\end{tabular}
\end{center}
\caption{\label{tab-res2}
An experimental comparison of the performance of Euclidean nearest
neighbor methods with statistical similarity based nearest neighbor
methods on a rapid writer adaptation problem.
The error rates are derived from 8767 test samples, using
10 prototypes selected as described in the paper.
}
\end{table}

\p

These experimental results demonstrate that using statistical
similarity functions can result in greatly improved recognition rates
compared to Euclidean nearest neighbor classification
methods--statistical similarity functions are an effective ``adaptive
metric'' for these kinds of problems.

However, that is all these initial experiments were designed
to test, and several important experiments remain to be done;
we will return to this issue in the Discussion.

\section{Learning Single View Generalization}

As a second problem to be addressed using statistical similarity,
we consider the problem of generalizing the appearance of a 3D
object to novel viewpoints given a single view.

As an example of statistical similarity, this is an interesting
problem because the problem cannot be solved as a classification
problem.

The problem occurs in both psychophysics and computer vision and
previous approaches to it in the literature have postulated
very specific representations of views and 3D objects in order
to admit such generalization;

a statistical similarity approach like the one presented here also
yields a simpler and potentially more general theory of such
single view generalization phenomena.

\p

In the following, we consider the space of 3D paperclips, as is
frequently used in both psychophysical experiments and theoretical
work on learning in computer vision \cite{PogEde90}.

That is, a model $M$ is an ordered set of points in $\R^3$.

Models are constructed by concatenating $k=5$ unit vectors in $\R^3$
at randomly chosen orientations, meaning that $M$ can be represented
as a vector in $\R^{15}$.

The method in which models $M$ are constructed randomly gives
rise to a prior distribution $P(M)$ over the space of models.

\p

Given a set of viewing parameters, $V$, we
derive a view $V$ from the model through a parameterized
imaging transformation $B = f_V(M)$;

here, the imaging transformation is assumed to be a rigid body
transformation followed by orthographic projection.

After the imaging transformation, the image $B$ is represented in one
of three different ways: as a list of $(x,y)$ feature locations in the
image (in vertex order), as list of 2D angles between successive
edges, and as a quantized features map on a $40\times 40$ grid, with
each grid square indicating the presence or absence of a vertex
within that square.

These are representations that have been commonly used in
experiments on 3D recognition on paperclips by previous authors
\cite{PogEde90}.

\p

The viewing transformation can be written as a
conditional density (this is simply expressing the same
functional relationship using the notation of a conditional
probability):

\begin{equation}
P(B|V,M) = \delta(B,f_V(M))
\end{equation}

Here, $\delta$ is the Dirac delta function.

In the presence of noise on the location of feature points or the
location of model points, the delta function is replaced by another
distribution related to the noise.

For example, under iid additive noise distributed according to a
distribution $N(x)$, the conditional distribution becomes:

\begin{equation}
P(B|V,M) = N(B-f_V(M))
\end{equation}

If we integrate out the (unobservable) distributions over noise and
viewing parameters, we are left with a marginal distribution $P(B|M)$,
the distribution of views of an model under these viewing conditions.

\p

The viewing parameters were represented by slant and tilt
(rotations around two axes perpendicular to the optical axis
of the observer).

Slant and tilt angles were either drawn uniformly randomly from the
interval $[-40^o,+40^o]$ or from the set $\{-45^o,+45^o\}$, giving a
prior distribution over viewing parameters, $P(V)$.

\p

Because of the projection involved in the imaging transform, there is
potentially an infinity of models that could have given rise to a
given image $B$.

For example, all models that differ only by their placement of
vertices along the optical axis after rigid body transformation and
the addition of noise are indistinguishable from their images.

\p

We use a forced choice framework of recognition.

In the simplest case, the observer is presented with two views and has
to decide whether they derive from the same 3D model or not.

This kind of visual object recognition problem occurs, for example, in
face verification, where an observer needs to compare two photographs
of faces and perform 3D generalization based on a single view.

A slightly more complicated forced choice problem is one in which an
observer is presented with a target view and two unknown views, one of
which is known to be derived from the same model as the target view
(condition $S=1$), and the other of which is derived from some other
randomly chosen model (condition $S=0$).

We might call this the ``police lineup'' problem, in which an observer
has to pick out a previously seen instance from a lineup known to
contain an instance.

These and similar forced choice experiments are commonly used in
psychophysical experiments on visual object recognition.

They have in common that there is no task-relevant 
classification or categorization of objects--two views derived from
two different models in one experiment may well come from the same
model in another trial.

\begin{figure}
\hbox{%
\fbox{\includegraphics[height=1in]{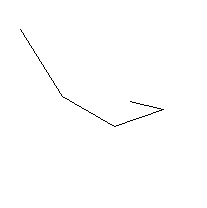}}~~~%
\fbox{\includegraphics[height=1in]{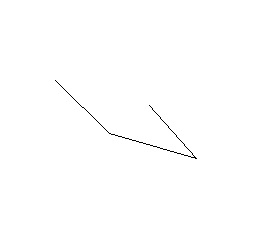}}~%
\fbox{\includegraphics[height=1in]{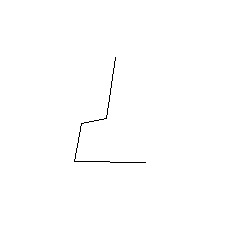}}%
}
\caption{\label{fig-noclass} An instance of forced choice used
in the 3D recognition experiments.}
\end{figure}

A representative instance of such a force choice problem is shown in
Figure~\ref{fig-noclass}.

\p

Let the target view be $B$ and $B'$ be one of the unknown views.

For concreteness, let us write down the relationship between $P(S|B,B')$
and the generative model expressed as $P(B|M)$.

In analogy to Equation~\ref{eqsame}, for $S=1$, we can expand in
terms of $P(M|B)$ and apply Bayes rule:

\begin{eqnarray}
P(S=1|B,B') = \int P(M|B) P(M|B') \,dM &&\\
= \int \frac{P(B|M) P(B'|M) P(M) P(M)}{P(B)P(B')} \,dM &&
\end{eqnarray}

Modeling and evaluating this integral and the individual factors
would be a difficult task;

even the much simpler problem of finding the maximum
a posteriori model matching from a fixed set of models,
that is, $\arg\max_{M\in\{M_1,\ldots,M_k\}} P(M|B)$, has
proved to be a daunting computational task (c.f.
\cite{wells97}).

\p

However, as in the character recognition example above, we can model
$P(S=1|B,B')$ directly using non-parametric models of probability
distributions.

To do this, we can generate images from various models under different
viewing parameters and train the probability model using images
derived from the same object model as positive examples and images
derived from different object models as negative training examples.

As before, a multi-layer perceptron (MLP) was used to model posterior
probability distributions.

\p

During training, a fixed set of 200 models was drawn once from the
distribution $P(M)$ and used for generating images.

During testing, new, previously unseen models were drawn from $P(M)$
and images generated from them as described above for the forced
choice framework.

The performance of statistical similarity was compared with the performance
of Euclidean distance in feature space (equivalent to a least square match
of the images when vertex locations are used as features).

The unknown view closest to the target according to statistical
similarity or Euclidean distance was returned as the view more likely
to have been derived from the same model as the target (when using
statistical similarity, this is easily seen to be the Bayes optimal
decision rule).

The results demonstrate an improvement in recognition performance
from a factor 1.8 to a factor 22 of statistical similarity compared
to nearest neighbor classification.

Note that in both cases, the generalization to arbitrary 3D paperclip
models was based on a limited training sample of only 200 paperclips.

\begin{table}[t]
\begin{tabular}{|p{2in}|p{.4in}|p{.4in}|} \hline
{\bf Features and Model} & {\bf Stat.} & {\bf Eucl.} \\
& {\bf Sim.} & {\bf Dist.} \\ \hline \hline
ordered angles MLP (8:100:1) &  10.9\% & 19.9\% \\ \hline
ordered locations, MLP (20:100:1) & 0.12\% & 0.86\% \\\hline
ordered locations, MLP (20:100:1), $\pm45^\circ$& ~0.38\% & ~8.4\% \\\hline
feature map, MLP (3200:100:1) & ~7.9\% & 32\% \\\hline
\end{tabular}
\caption{
Experiments evaluating MLP-based statistical 
similarity relative to view based recognition using 2D similarity.
Error rates (in percent) achieved by MLP-based statistical view
similarity models relative to error rates based on Euclidean distance
(equivalent to 2D similarity in the case of location features).  In
all experiments, the training set consisted of 200 clips consisting
each of five vertices.  The test set consisted of 10000 previously
unseen clips drawn from the same distribution.  The structure of the
network is given as ``($n$:$m$:$r$)'', where $n$ is the number of
inputs, $m$ the number of hidden units, and $r$ the number of outputs.
\label{tab-results}
}
\end{table}

\section{Discussion}

This paper has introduced the notion of statistical similarity based
on the conditional distribution $P(S=1|x,x') = \sum_\omega P(\omega|x)
P(\omega|x')$.

It was shown that classification using a statistical similarity
function and a set of unambiguous exemplars is equivalent to Bayesian
minimum error rate classification.

The paper has also brought variable metric nearest neighbor methods
into the framework of statistical similarity measures.

The paper has presented two sets of experiments.

\p

First, it has compared the performance of statistical similarity
with a Euclidean nearest neighbor classifier on two handwritten
character recognition problems.

Those experiments demonstrated a significant improvement relative
to Euclidean nearest neighbor methods.

This result should be considered merely a ``sanity check''--it shows
that the method can be used to construct similarity measures that are
significantly better than the baseline of Euclidean distance.

Whether using statistical similarity as an improved ``distance'' in a
nearest neighbor classifier ultimately will result in a
state-of-the-art character recognition system (e.g., \cite{LeCun98})
is not clearly addressed by these results.

A straightforward application of $k$-nearest neighbor classification
using statistical similarity (or Euclidean distances, for that matter)
is impractical anyway because it is too slow.

However, the statistical similarity measure introduced in this paper
can easily be used as part of a hierarchical or partitioned nearest
neighbor classifier, and this is likely to be the best route towards
constructing a statistical similarity based classifier that can handle
the large number of prototypes needed to achieve state of the art
performance.

This remains to be done for future work.

Of course, another set of experiments that would be desirable would be
direct comparisons on the same dataset with adaptive nearest neighbor
methods like those described by
\cite{Lowe95,Friedman94,HasTib96,shankar00weight,DomPenGun00}.

\p

A second set of experiments compared the performance of statistical
similarity with the performance of Euclidean nearest methods on a 3D
generalization problem in visual object recognition.

This example is interesting because it lacks a class structure;
as shown in \cite{CleJac91}, it is impossible to partition a set of
3D models into non-overlapping sets of views.

In this case, similarity is not a means to an end, as in nearest
neighbor classifiers, but it is an essential component of the problem--the
system really needs to be able to carry out well-founded
similarity judgements among objects in order to perform well.

The experiments on single view generalization using statistical
similarity show that it gives greatly improved performance relative to
2D similarity.

A practical advantage of the statistical similarity approach to single
view generalization compared to previous approaches
\cite{RiePog99,RiePog00,DuvEde99} is that it does not need to
postulate any kind of special problem structure (hierarchical feature
extraction, interpolable prototypes, class membership).

\p

Overall, this paper has outlined the beginnings of a Bayesian theory
of learning similarity.

As we noted in the introduction, statistical similarity is already
implicitly making an appearance in a number of areas of computer
vision, pattern recognition, and information retrieval.

Perhaps its most important contribution is to show that notions of
similarity that have previously been discussed in the form of
geometrically motivated ``distance measures'' or that are based on
dyadic probability models having a specific parametric forms can be
understood in, and unified under, a general Bayesian view.

\p

In the future, it will be important to see whether other forms
of statistical similarities may be easier to estimate or
manipulate; for example the conditional distribution $P(B'|B,S)$
is an alternative to $P(S=1|B,B')$ and has some computational
advantages.

While the framework of statistical similarity allows us to plug
in arbitrary classifiers and features, some classifiers and
feature types may turn out to be better suited to these kinds
of problems.

It has taken many years for the community to gain experience
with this in the context of traditional classification problems,
and it will likely take some time to gain similar experience for
statistical similarity.

As part of this, much more extensive benchmarking and
performance evaluations than could be presented here will
be needed.

Some of the most promising applications of statistical similarity
are on problems where existing classification-based approaches
don't apply at all (e.g., face verification, some information
retrieval problems), or where rapid adaptation to novel
styles or problems are needed (e.g., multi-font OCR,
on-line handwriting recognition).

\bibliographystyle{plain}
\bibliography{recog}

\end{document}